\documentclass[10pt,twocolumn,letterpaper]{article}

\usepackage{iccv}
\usepackage{times}
\usepackage{epsfig}
\usepackage{graphicx}
\usepackage{amsmath}
\usepackage{amssymb}
\usepackage{booktabs}
\usepackage{multirow}
\usepackage[accsupp]{axessibility}

\usepackage[breaklinks=true,bookmarks=false]{hyperref}

\iccvfinalcopy % *** Uncomment this line for the final submission

\def\iccvPaperID{4302} % *** Enter the ICCV Paper ID here

\ificcvfinal\fi

\begin{document}

%%%%%%%%% TITLE
\title{Dynamic Snake Convolution based on Topological Geometric Constraints for Tubular Structure Segmentation}

\author{Yaolei Qi$^{1}$, Yuting He$^{1}$, Xiaoming Qi$^{1}$, Yuan Zhang$^{1}$, and Guanyu Yang$^{1,2,3}$\thanks{Corresponding author}\\
$^{1}$ Key Laboratory of New Generation Artificial Intelligence Technology and Its\\
Interdisciplinary Applications (Southeast University), Ministry of Education, Nanjing 210096, China\\
$^{2}$ Jiangsu Province Joint International Research Laboratory of Medical Information Processing,\\ 
Southeast University, Nanjing, China\\
$^{3}$ Centre de Recherche en Information Biomédicale Sino-Fran${\c{c}}$ais (CRIBs),
Strasbourg, France\\
{\tt\small yaolei710@seu.edu.cn, yang.list@seu.edu.cn}
}

\maketitle
\ificcvfinal\thispagestyle{empty}\fi

%%%%%%%%% ABSTRACT
\begin{abstract}
   Accurate segmentation of topological tubular structures, such as blood vessels and roads, is crucial in various fields, ensuring accuracy and efficiency in downstream tasks. However, many factors complicate the task, including thin local structures and variable global morphologies. In this work, we note the specificity of tubular structures and use this knowledge to guide our DSCNet to simultaneously enhance perception in three stages: feature extraction, feature fusion, and loss constraint. First, we propose a dynamic snake convolution to accurately capture the features of tubular structures by adaptively focusing on slender and tortuous local structures. Subsequently, we propose a multi-view feature fusion strategy to complement the attention to features from multiple perspectives during feature fusion, ensuring the retention of important information from different global morphologies. Finally, a continuity constraint loss function, based on persistent homology, is proposed to constrain the topological continuity of the segmentation better. Experiments on 2D and 3D datasets show that our DSCNet provides better accuracy and continuity on the tubular structure segmentation task compared with several methods. Our codes are publicly available\footnote{\url{https://github.com/YaoleiQi/DSCNet}}.
\end{abstract}

%%%%%%%%% BODY TEXT
\section{Introduction}

%--------------------------------Background-------------------------------
The accurate segmentation of topological tubular structures is paramount in various fields to ensure the precision and efficiency of downstream tasks. In clinical applications, a well-delineated blood vessel is a crucial prerequisite for computational hemodynamics, and it assists radiologists in locating and diagnosing lesions \cite{Intro01KoJACC, Intro02MinJAMA}. In remote sensing applications, complete road segmentation provides a solid foundation for route planning. Regardless of the field, these structures share common features of being thin and tortuous, which make them challenging to capture due to their small proportion in images. Therefore, there is an urgent need to enhance the perception of thin tubular structures.

%--------------------------------Challenges------------------------------
\begin{figure}[t]
  \centering
%  \fbox{\rule{0pt}{2in} \rule{0.9\linewidth}{0pt}}
   \includegraphics[width=\linewidth]{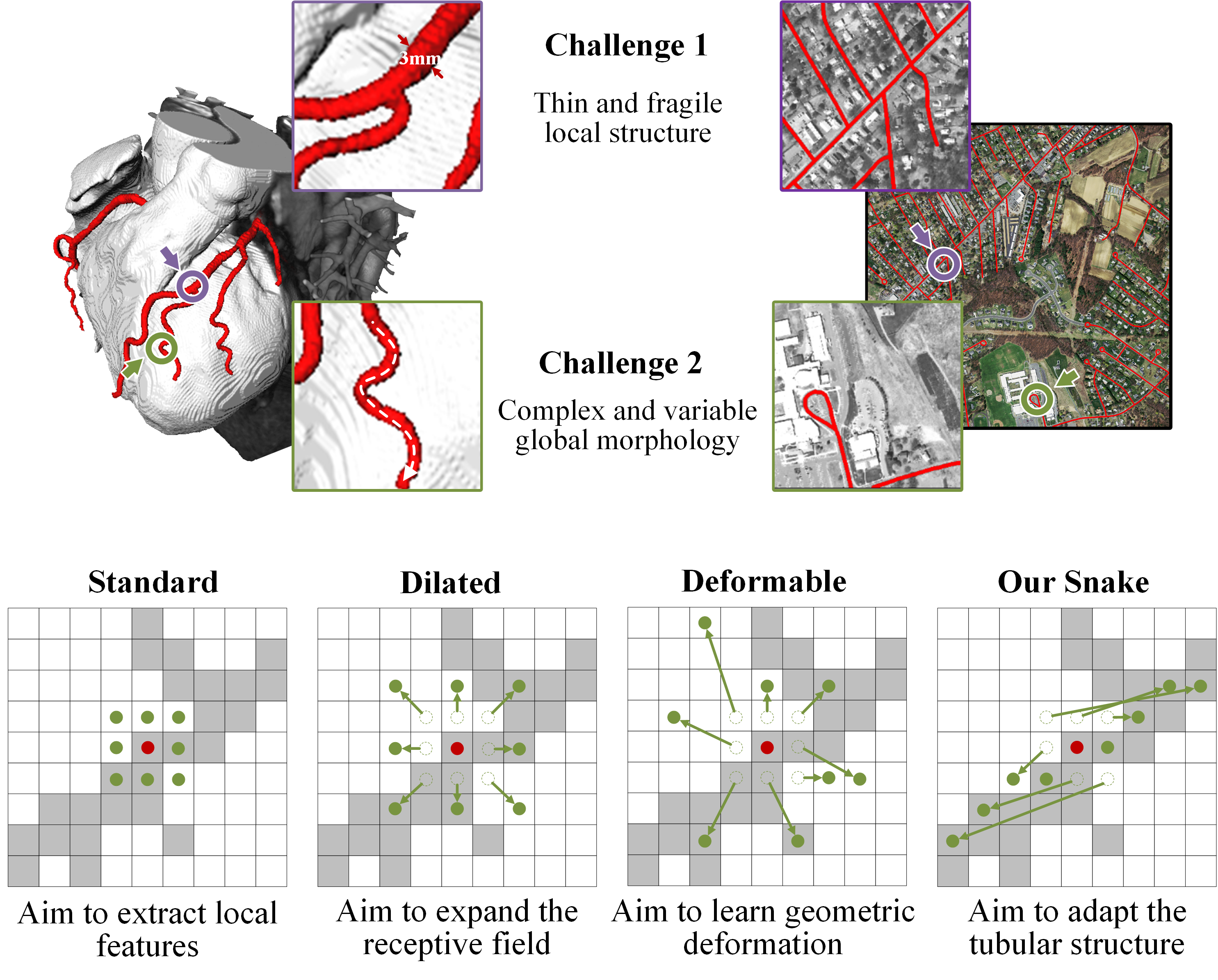}
   \caption{
   \textbf{Challenges.} The above figure shows a 3D heart vascular dataset and a 2D remote road dataset. Both datasets aim to extract tubular structures, but this task faces challenges due to fragile local structures and complex global morphology. \textbf{Motivation.} The standard convolutional kernel is intended to extract local features. On this basis, deformable convolutional kernels have been designed to enrich their application and adapt to geometric deformations of different targets. However, due to the aforementioned challenges, it is difficult to focus efficiently on the thin tubular structures.
   }
   \label{fig:Inro_one}
\end{figure}

However, it remains challenging due to the following difficulties: (1) \textbf{Thin and fragile local structure.} As shown in Figure~\ref{fig:Inro_one}, thin structures account for only a small proportion of the overall image with limited pixel composition. Moreover, these structures are susceptible to interference from complex backgrounds, rendering it difficult to precisely discriminate subtle target variations by the model. Consequently, the model may struggle to differentiate these structures, resulting in the fracture of the segmentation. (2) \textbf{Complex and variable global morphology.} Figure~\ref{fig:Inro_one} shows the complex and variable morphology of thin tubular structures, even within the same image. Morphological variations are observed in targets located in different regions, depending on the number of branches, the location of bifurcations, and the path length. The model may tend to overfit features that have already been seen, resulting in weak generalization when the data exhibits unprecedented morphological structures.

%------------------------------Related_Work------------------------------
%according to the stage of knowledge usage:% 
Recently, many studies have proposed incorporating domain knowledge (\eg geometric topology and tree structure) to guide the model better to perceive the distinctive features of the tubular structures, thus focusing on improving the accuracy of local segmentation and maintaining the continuity of global morphology. Existing methods can be broadly classified into three categories: (1) Network-based methods \cite{RW_DCN_CMR, RW_DCN_retinal1, RW_DCN_retinal2, RW_Network_Points, RW_Network_TreeLSTM, RW_Net_wavelet} design specific network architectures according to the characteristics of the tubular structures that guide the model to focus on critical features. However, given the small proportion of tubular structures, the network may inevitably lose the perception of the corresponding structures. (2) Feature-based methods \cite{RW_FF_CMR, RW_FF_MICCAI, RW_FF_global_local, RW_FF_graph, RW_FF_mvsgan} enhance the comprehension of the particular geometric and topological features of tubular structures by supplementing the model with additional feature representations. However, some redundant feature representations exacerbate the computational burden while not providing a positive influence on the model. (3) Loss-based methods \cite{RW_Loss_clDice, RW_Loss_DDT, RW_Loss_Topological, RW_Loss_PH} incorporate measurement methods to supplement constraints during the training process, typically through loss functions. These methods reinforce the stringent constraints on segmentation. Building on this foundation, structured losses combined with continuity constraints from the topological perspective will potentially further improve the accuracy of the tubular segmentation.

%-------------------------------Motivation-------------------------------

To tackle the above obstacles, we propose a novel framework, DSCNet, involving a tubular-aware convolution kernel, multi-view feature fusion strategy, and topological continuity constraint loss function. (1) To address the challenge of the small proportion of thin and fragile local structures that are difficult for the model to focus on, we propose Dynamic Serpentine Convolution (DSConv) to enhance the perception of the geometric structure by adaptively focusing on the thin and curved local features of tubular structures. Unlike deformable convolution \cite{RW_DCN}, which leaves the network completely free to learn geometrical changes, thus causing perceptual regions to wander, especially on thin tubular structures, our DSConv considers the snake-like morphology of tubular structures and supplements the free learning process with constraints that allow for targeted enhancement of the perception of tubular structures. (2) To address the challenge of complex and variable global morphology, we propose a multi-view feature fusion strategy. In this method, we generate multiple morphological kernel templates based on DSConv for viewing the structural features of the target from various perspectives and achieve efficient feature fusion by summarizing typical vital features. (3) To address the problem that segmentation of tubular structures is prone to fracture, we propose a Topological Continuity Constrained Loss Function (TCLoss) based on Persistent Homology (PH). PH\cite{PH_1, PH_2, PH_3} responds to the process of topological features from emergence to disappearance. It obtains adequate topological information from noisy high-dimensional data. The related Betti numbers are one way of describing connectivity in the topological space. Unlike \cite{RW_Loss_PH, PH_WD}, our TCLoss combines PH with point set similarity to pilot the network to focus on fracture regions with anomalous pixel/voxel distributions, achieving continuity constraints from a topological perspective.

%------------------------------Contribution------------------------------

To summarize, our work proposes a novel framework of knowledge fusion addressing the difficulties of the thin tubular structures, and the specific contributions are three-fold. (1) We propose a dynamic snake convolution to adaptively focus on the slender and tortuous local features and realize the accurate tubular structures segmentation on both 2D and 3D datasets. Our model is thoroughly verified using both internal and external test data. (2) We propose a multi-perspective feature fusion strategy to supplement the attention to the vital features from multiple perspectives. (3) We propose a topological continuity constraint loss function based on Persistent Homology, which better constrains the continuity of the segmentation. 

%-----------------------------------------------------------------
\section{Related Work}
\subsection{Methods based on Network Design}
Various methods have been proposed to achieve better performance by designing particular network architectures and modules according to the morphology of the tubular structures. (1) Methods based on the convolutional kernel design, represented by the famous dilated convolution \cite{RW_Dilated} and deformable convolution \cite{RW_DCN}, are proposed to deal with the inherent limited geometric transformation in CNNs, showing outstanding performance on sophisticated detection and segmentation tasks. These methods \cite{RW_DCN_CMR, RW_TIP_superpixel, RW_DCN_retinal1, RW_DCN_retinal2} are also designed to dynamically perceive the geometric features of the objects to adapt to the structure with changeable morphology. For example, DUNet proposed in \cite{RW_DCN_retinal1} integrates the deformable convolution into the U-shape architecture and adaptively adjusts the receptive field according to the vessels' scales and shapes. (2) Methods based on the network architecture design are proposed to learn the special geometric topological features of the tubular structures. PointScatter \cite{RW_Network_Points} is proposed to represent the tubular structure with points set, which is an alternative to the segmentation models for the tubular structure extraction task. \cite{RW_Network_TreeLSTM} proposed a tree-structured convolutional gated recurrent unit to model the coronary artery's topological structures explicitly. Different from the ideas mentioned above of allowing the model to learn the geometric changes completely freely, considering the limitation of the convergence difficulty caused by excessive randomness and the possibility that the model will focus on the unexpected regions of the target. Our work integrates the domain knowledge of tubular structure morphology to stably enhance the perception of the tubular structure in the feature extraction process.

%-----------------------------------------------------------------
\subsection{Methods based on Features Fusion}
Methods based on features fusion \cite{RW_FF_graph, RW_FF_global_local, RW_FF_MICCAI, RW_FF_CMR, RW_FF_mvsgan} strengthens the representation of the tubular structure by supplementing additional feature information to the model. Considering the topology and the sparsity of the tubular structures, \cite{RW_FF_graph} proposed a cross-network multi-scale feature fusion method performed between two networks to support high-quality vessel segmentation effectively. In \cite{RW_FF_global_local}, a global transformer and dual local attention network via deep-shallow hierarchical feature fusion are investigated to simultaneously capture the global and local characterizations. \cite{RW_FF_MICCAI} proposed to fuse the contextual anatomical information and vascular topologies for accurate tubular structure segmentation. In our work, we propose a multi-perspective feature fusion strategy to supplement the attention to the vital features from multiple perspectives. In this strategy, we generate numerous morphological kernel templates based on our DSConv to observe the structural characteristics of the target from multiple perspectives and realize the feature fusion by summarizing the essential standard features, thus improving the performance of our model.

%-----------------------------------------------------------------
\subsection{Methods based on Loss Function}
Methods based on loss function \cite{RW_Loss_clDice, RW_Loss_DDT, RW_Loss_Topological} introduce measurement methods to supplement constraints in the training process. These methods strengthen the strong constraints on the tubular structures segmentation. \cite{RW_Loss_clDice} introduced a similarity measurement termed centerline Dice, which is calculated on the intersection of the segmentation masks and the skeleton. \cite{RW_Loss_DDT} proposed a geometry-aware tubular structure segmentation method, Deep Distance Transform (DDT), which combines intuitions from the classical distance transform for skeletonization and tubular structure segmentation. These methods focus on the continuity of the tubular structure segmentation, but the skeleton's inaccuracy and offset will affect the constraints' precision. \cite{RW_Loss_Topological} proposed a similarity index that captures the topological consistency of the predicted segmentation and designs a loss function based on the morphological closing operator for tubular structure segmentation. In \cite{RW_Loss_PH}, topological data analysis methods are incorporated with a geometric deep learning model for fine-grained segmentation for 3D objects. These methods will capture the features of the topological objects. Drawing inspiration from this, our work proposes a topological continuity constraint loss function (TCLoss) that better constrains the continuity of the segmentation from a topological perspective. Our TCLoss gradually introduces constraints based on Persistence Homology \cite{RISPER1, RISPER2} during the training process to guide the network to focus on the fracture regions and realize continuity. 

\begin{figure*}[t]
  \centering
%  \fbox{\rule{0pt}{2in} \rule{0.9\linewidth}{0pt}}
   \includegraphics[width=\linewidth]{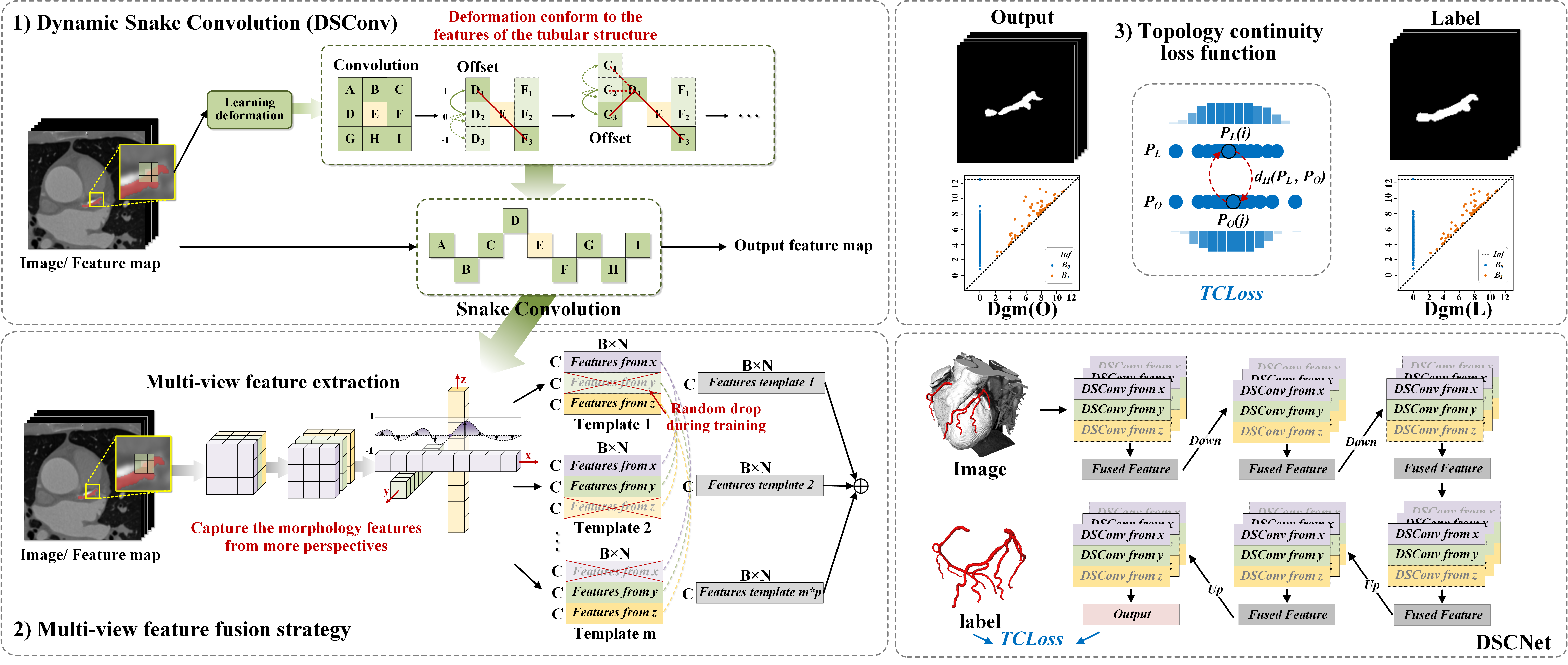}
   \caption{
   \textbf{Methodology.} Schematic overview of our proposed method illustrated on an example of the 3D coronary artery segmentation. Our method has three sections: (1) Dynamic snake convolution (DSConv), which learns the deformation according to the input feature map, adaptively focuses on the slender and tortuous local features under the knowledge of the tubular structure morphology. (2) Multi-view feature fusion strategy, which generates multiple morphological kernel templates based on our DSConv and is used to observe the structural characteristics of the target from multiple perspectives. (3) Loss function, called topological continuity constraint loss function (TCLoss), is based on Persistent Homology to guide the network to focus on the fracture regions with abnormally low pixels/voxels distribution and realize continuity constraint.
   }
   \label{fig:Method_framework}
\end{figure*}

\section{Methodology}
\label{sec:method}

Our method is designed to deal simultaneously with 2D and 3D feature maps of thin tubular structures. For simplicity, our modules are described in 2D, and the detailed extension to 3D is also provided in our open source.

%-----------------------------------------------------------------
\subsection{Dynamic Snake Convolution}

\begin{figure}[b]
  \centering
%  \fbox{\rule{0pt}{2in} \rule{0.9\linewidth}{0pt}}
   \includegraphics[width=\linewidth]{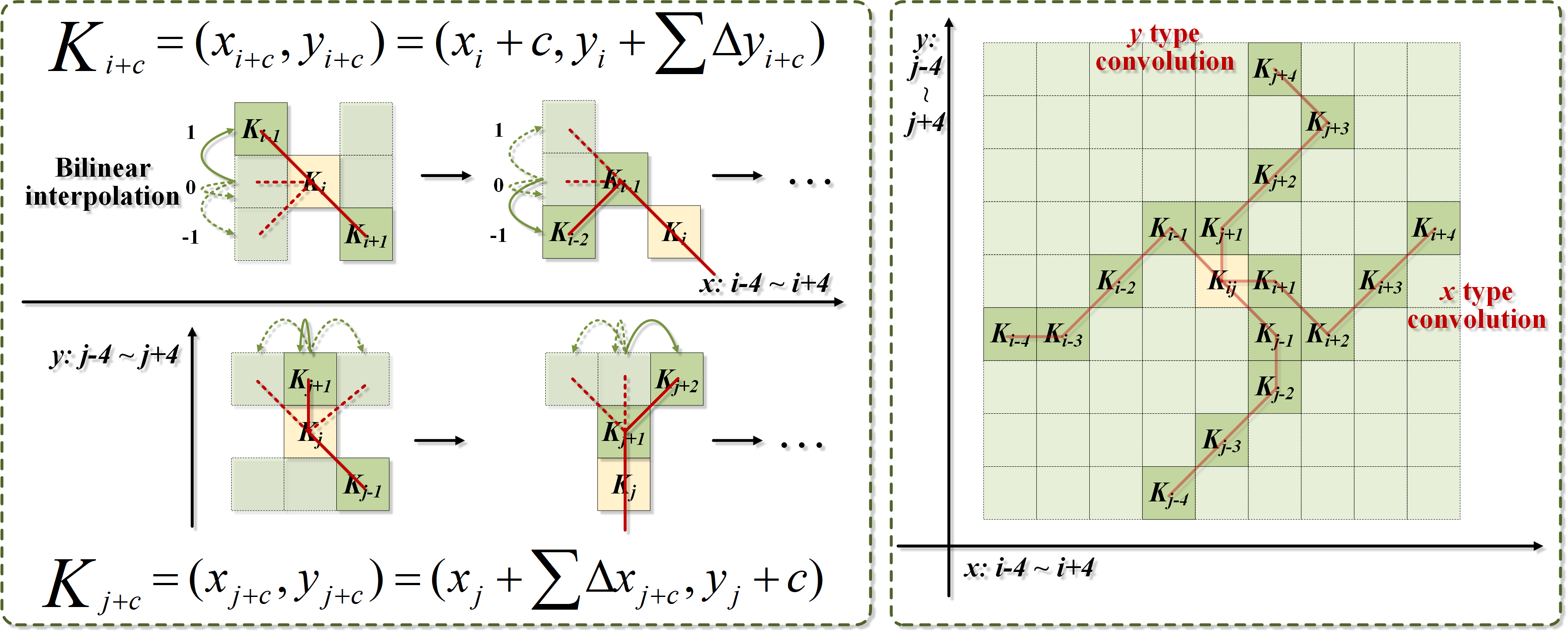}
   \caption{
   \textbf{Left:} Illustration of the coordinates calculation of the DSConv. \textbf{Right:} The receptive field of the DSConv.
   }
   \label{fig:Method_one}
\end{figure}

In this section, we discuss how to perform Dynamic Snake Convolution (DSConv) to extract the tubular structure's local features. Given the standard 2D convolution coordinates as $K$, the central coordinate is $K_i=(x_i, y_i)$. A $3\times3$ kernel $K$ with dilation $1$ is expressed as:
\begin{equation}
	K=\{(x-1, y-1), (x-1, y), \cdots, (x+1, y+1)\}
	\label{eq:2Dconv_coord}
\end{equation}

To give the convolution kernel more flexibility to focus on complex geometric features of the target, inspired by \cite{RW_DCN}, we introduced deformation offsets $\Delta$. However, if the model is left free to learn the deformation offsets, the perceptual field tends to stray outside the target, especially in the case of thin tubular structures. Therefore, we use an iterative strategy (Figure~\ref{fig:Method_one}), selecting the following position to be observed in turn for each target to be processed, thus ensuring continuity of the attention and not spreading the field of sensation too far due to the large deformation offsets. 

In DSConv, we straighten the standard convolution kernel, both in the direction of the x-axis and the y-axis. We consider a convolution kernel of size 9 and take the x-axis direction as an example, the specific position of each grid in $K$ is represented as: $K_{i\pm c}=(x_{i \pm c}, y_{i \pm c})$, where $c=\{0, 1, 2, 3, 4\}$ denotes the horizontal distance from the central grid. The selection of each grid position $K_{i\pm c}$ in the convolution kernel $K$ is a cumulative process. Starting from the center position $K_{i}$, the position away from the center grid depends on the position of the previous grid: $K_{i+1}$ is augmented with an offset $\Delta=\{\delta|\delta \in [-1, 1]\}$ compared to $K_{i}$. Hence, the offset needs to be $\Sigma$, thus ensuring that the convolution kernel conforms to a linear morphological structure. Figure~\ref{fig:Method_one} in the direction of the x-axis becomes:

\begin{equation}
K_{i\pm c}=\left\{
    \begin{aligned}
    (x_{i+c}, y_{i+c}) = (x_i+c, y_i + \Sigma_{i}^{i+c} \Delta y), \\
    (x_{i-c}, y_{i-c}) = (x_i-c, y_i + \Sigma_{i-c}^{i} \Delta y), \\
    \end{aligned}
    \right. 
	\label{eq:2DDSConv_coord_x}
\end{equation}

and Equation~\ref{eq:2DDSConv_coord_x} in the direction of the y-axis becomes:

\begin{equation}
K_{j\pm c}=\left\{
    \begin{aligned}
    (x_{j+c}, y_{j+c}) = (x_{j} + \Sigma_{j}^{j+c} \Delta x, y_j+c), \\
    (x_{j-c}, y_{j-c}) = (x_{j} + \Sigma_{j-c}^{j} \Delta x, y_j-c), \\
    \end{aligned}
    \right. 
	\label{eq:2DDSConv_coord_y}
\end{equation}

As the offset $\Delta$ is typically fractional, bilinear interpolation is implemented as:

\begin{equation}
K = \Sigma_{K'} B(K', K)\cdot K'
	\label{eq:2DDSConv_bilinear_interpolation}
\end{equation}

where $K$ denotes a fractional location for Equation~\ref{eq:2DDSConv_coord_x} and Equation~\ref{eq:2DDSConv_coord_y}, $K'$ enumerates all integral spatial locations and $B$ is the bilinear interpolation kernel and it is separated into two one-dimensional kernels as: 

\begin{equation}
B(K, K') = b(K_x,K'_x)\cdot b(K_y,K'_y)
	\label{eq:bilinear}
\end{equation}

As shown in Figure~\ref{fig:Method_one}, our DSConv covers a $9 \times 9$ range during the deformation process due to the two-dimensional (x-axis, y-axis) changes. DSConv is designed to better adapt to the slender tubular structure based on the dynamic structures so as to better perceive the key features. 

%-------------------------------------------------------------------------
\subsection{Multi-view Feature Fusion Strategy}

This section discusses implementing the multi-view feature fusion strategy to guide the model to complement the focus on essential features from multiple perspectives. For each $K$, two feature maps $f^{l}(K_x)$ and $f^{l}(K_y)$ from layer $l$ are extracted from the x-axis and y-axis, expressed as:

\begin{equation}
f^{l}(K)=\{\underbrace{\Sigma_i w(K_i) \cdot f^l(K_i)}_{f^{l}(K_x)}, 
\underbrace{\Sigma_j w(K_j)\cdot f^l(K_j)}_{f^{l}(K_y)}\}
	\label{eq:multi_1}
\end{equation}

where $w(K_i)$ denotes the weight at position $K_i$, and the features extracted by the $l$-th layer convolution kernel $K$ are calculated using the cumulative approach.

Based on Equation~\ref{eq:multi_1}, we extract $m$ groups of features as $T^l$, which contains different morphology of the DSConv:

\begin{equation}
    T^l = (
                \underbrace{f^l(K_x), f^l(K_y)}_{T^l_1},
                \underbrace{f^l(K_x), f^l(K_y)}_{T^l_2},
                \cdots
                \underbrace{f^l(K_x), f^l(K_y)}_{T^l_m}
            )
   \label{eq:multi_2}
\end{equation}

\begin{figure}[t]
  \centering
%  \fbox{\rule{0pt}{2in} \rule{0.9\linewidth}{0pt}}
   \includegraphics[width=\linewidth]{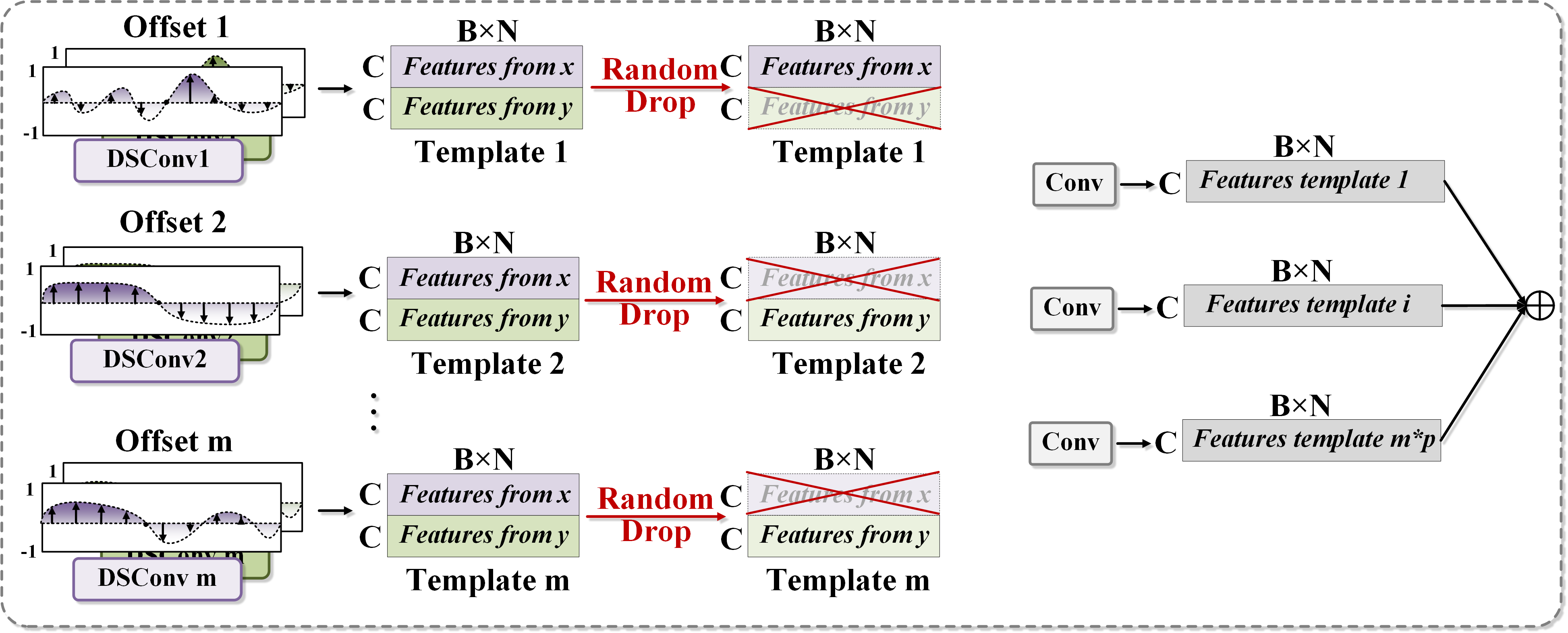}
   \caption{
   The multi-view feature fusion strategy.
   }
   \label{fig:Method_two}
\end{figure}

The feature fusion of multiple templates will inevitably bring redundant noise. Therefore, a random dropping strategy $r^l$ is introduced during the training stage (Figure~\ref{fig:Method_two}), to improve the performance of our model and prevent over-fitting without increasing additional computational burden, then Equation~\ref{eq:multi_2} becomes:

\begin{equation}   
\label{eq:multi_3}
    \left\{
    \begin{aligned}
        & r^l \sim \mbox{Bernoulli}(p)\\
        & \hat{T^l} = r^l \cdot T^l \\
        & f^{l+1}(K)= \Sigma^{\lfloor m\times p \rfloor}\hat{T^l_t} \\
    \end{aligned}
    \right.
\end{equation}

where $p$ is the probability of random dropping and $r^l$ satisfies the Bernoulli distribution. The optimal dropping strategy is saved during the training phase and guides the model to fuse key features during the testing phase.

%-------------------------------------------------------------------------
\subsection{Topological Continuity Constraint Loss}

In this section, we discuss how to implement the topological continuity constraint loss (TCLoss) based on the Persistent Homology to constrain the continuity of the segmentation. The geometric and topological information in complex structures is a pivotal clue to help the models understand the continuous structures. Tools from the topological data analysis are adopted to extract the essential features hidden in the complex tubular structures.

\begin{figure}[t]
  \centering
%  \fbox{\rule{0pt}{2in} \rule{0.9\linewidth}{0pt}}
   \includegraphics[width=\linewidth]{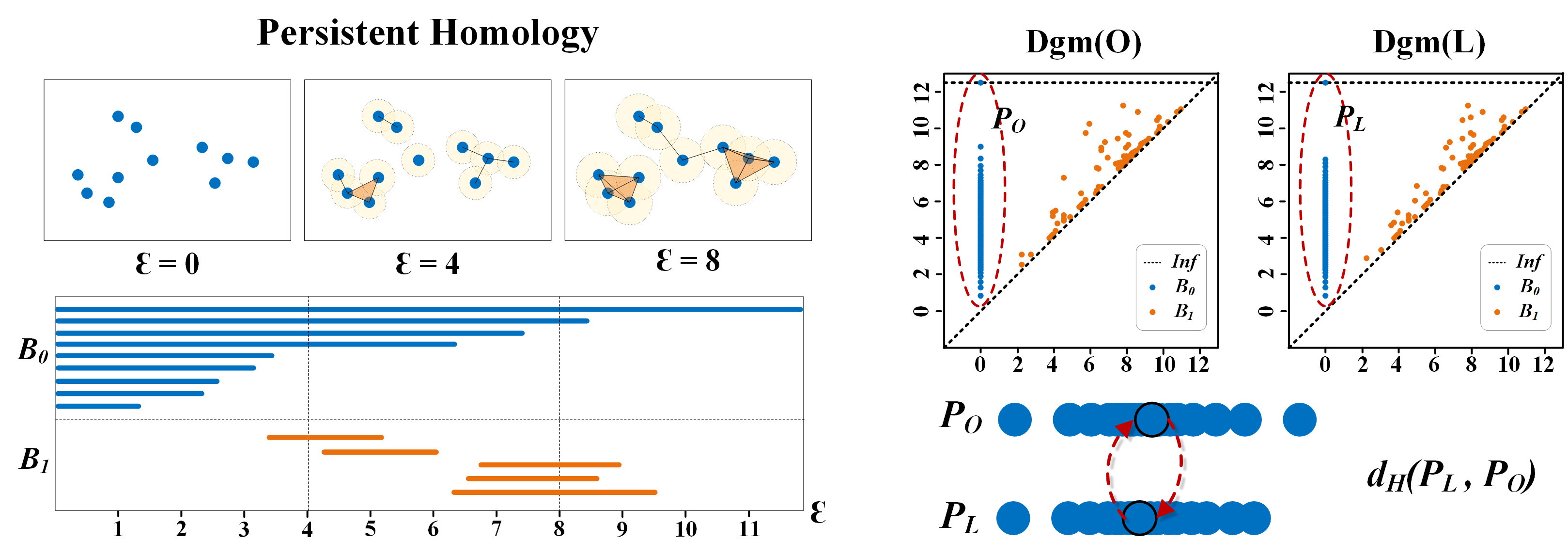}
   \caption{
   Illustration of the Persistent Homology and our TCLoss.
   }
   \label{fig:Method_three}
\end{figure}

We aim to construct the topology of the data and extract the high-dimensional relationships in the complex tubular structure, represented as persistence barcodes and Persistence Homology (PH), as illustrated in Figure~\ref{fig:Method_three}.

Given $G$, its $N$-dimensional topological structure, homology class \cite{PH_1, PH_2} is an equivalence class of $N$-manifolds which can be deformed into each other within $G$, where $0$-dimensional and $1$-dimensional are connected components and handles. PH is applied to compute the evolution of topological features, and the period between the appearance time $b$ and disappearance time $d$ of topological features is kept \cite{RW_Loss_PH}. Such periods are summarized in a concise format called a persistence diagram (PD), which consists of a set of points $(b, d)$. Each point $(b, d)$ represents the $d$-th homology class that appears at $b$ and disappears at $d$. Let $PD=\mbox{dgm}(\cdot)$ denote the persistent homology obtained from the groundtruth $L$ and the output $O$. We consider the topological information in complex tubular structures, which contains the key clues to determine the presence of fractures, to be evident in the homotopy features of $0$-dimensional and $1$-dimensional homological features. The existing methods \cite{RW_Loss_PH, PH_3, PH_WD} use modified Wasserstein distance to compute the best match between the points generated by the output and the points generated by the groundtruth, and the outlier points without optimal pairing are matched to the diagonal and do not participate in the loss calculation. However, in our task, the outlier points represent anomalous appearing or disappearing time and imply wrong topological relations that play an important role. Therefore, we measure the similarity between the two sets of points using Hausdorff distance\cite{HD}:

\begin{equation}   
\label{eq:TCLoss_1}
    \left\{
    \begin{aligned}
& d_H(P_O, P_L) = \max_{u \in P_O} \min_{v \in P_L} \parallel u - v \parallel \\
& d_H(P_L, P_O) = \max_{v \in P_L} \min_{u \in P_O} \parallel v - u \parallel\\
& d_H^*= \max \{ d_H(P_O, P_L),  d_H(P_L, P_O) \} \\
    \end{aligned}
    \right.
\end{equation}

where $P_O \in \mbox{Dgm}(O)$ , $P_L \in \mbox{Dgm}(L)$ and $d_H^*$ represents the bidirectional Hausdorff distance, which is computed in terms of $n$-dim points. Our used Hausdorff distance is sensitive to outliers. As shown in Equation~\ref{eq:TCLoss_1}, if two sets of points are similar, all the points are perfectly superimposed except only one point in $P_O$, which is far from any point in $P_H$, then the Hausdorff distance is determined by that point and is large \cite{Hausdorff_1}.

Then the summation over all dimensions ($n=0, 1, 2, \cdots, N$) is performed to obtain the $\mathcal{L}_{PH}$ and the whole TCLoss is integrated with the cross entropy loss $\mathcal{L}_{CE}$ as the final loss function $\mathcal{L}_{TC} = \mathcal{L}_{CE} + \sum_{n=0}^{N} d_H^*$.

%\begin{equation}
%\mathcal{L}_{TC} = \sum_{n=0}^{N} d_H^*
%	\label{eq:TCLoss2}
%\end{equation}

% \begin{equation}
% \mathcal{L}_{TC} = \mathcal{L}_{CE} + \mathcal{L}_{PH} =\mathcal{L}_{CE} + \sum_{n=0}^{N} d_H^*
% 	\label{eq:TCLoss3}
% \end{equation}

Finally, the topology and accuracy are constrained by the combined effect of the two loss functions, contributing to continuous tubular segmentation.

%%%%%%%%% EXPERIMENTS
%------------------------------------------------------------------------
\section{Experiments Configurations}
\label{sec:exp_config}

%------------------------------------------------------------------------
\subsection{Datasets}
We employ three datasets containing two public and one internal dataset for validating our framework. In 2D, we evaluate the DRIVE retina dataset\cite{Exp_vessel_2d} and the Massachusetts Roads dataset\cite{Exp_road_2D}. In 3D, we used a dataset called Cardiac CCTA Data. Details concerning the experimental setup can be found in the supplementary material.

\subsection{Evaluation Metrics}
We performed comparative experiments and ablation studies to demonstrate the advantages of our proposed framework. The classical segmentation network U-Net \cite{Exp_Unet} and the CS$^2$-Net \cite{Exp_CS2} proposed in 2021 for vascular segmentation are compared to validate the accuracy. To validate the network design performance, we compared the DCU-net \cite{RW_DCN_retinal2} proposed in 2022 for retinal vascular segmentation. To validate the advantages of feature fusion, we compared the Transunet \cite{Exp_transunet} for medical image segmentation proposed in 2021. To validate the loss function constraint, we compared clDice \cite{RW_Loss_clDice} proposed in 2021 and Wasserstein-distance-based TCLoss $\mathcal{L}_{WTC}$ \cite{RW_Loss_PH}. These models are trained on the same dataset with the exact implementation and were evaluated by the following metrics. All metrics were calculated for each image and averaged.

%DDT \cite{RW_Loss_DDT} proposed in 2020, 

\begin{enumerate}
	\item Volumetric scores: \textit{Mean Dice Coefficient (Dice)}, \textit{Relative-Dice coefficient (RDice)}\cite{RW_EE}, \textit{CenterlineDice (clDice)}\cite{RW_Loss_clDice}, Accuracy (ACC) and AUC are used to evaluate the performance of the results
	\item Topology errors: We follow \cite{RW_Loss_clDice, RW_Network_Points} and calculate the topology-based scores including the \textit{Betti Errors} for Betti numbers $\beta_0$ and $\beta_1$. Meanwhile, to objectively verify the continuity of the coronary artery segmentation, the \textit{overlap until first error (OF)} \cite{OF} is used to evaluate the completeness of the extracted centerline.
	\item Distance errors: \textit{Hausdorff Distance (HD)} \cite{HD} is also widely used to describe the similarity between two sets of points, which is recommended to evaluate the thin tubular structures.
	
	\end{enumerate}

%%%%%%%%% RESULTS
\section{Results and Discussion}
\label{sec:res}

%------------------------------------------------------------------------
\begin{figure*}
  \centering
%  \fbox{\rule{0pt}{2in} \rule{0.9\linewidth}{0pt}}
   \includegraphics[width=\linewidth]{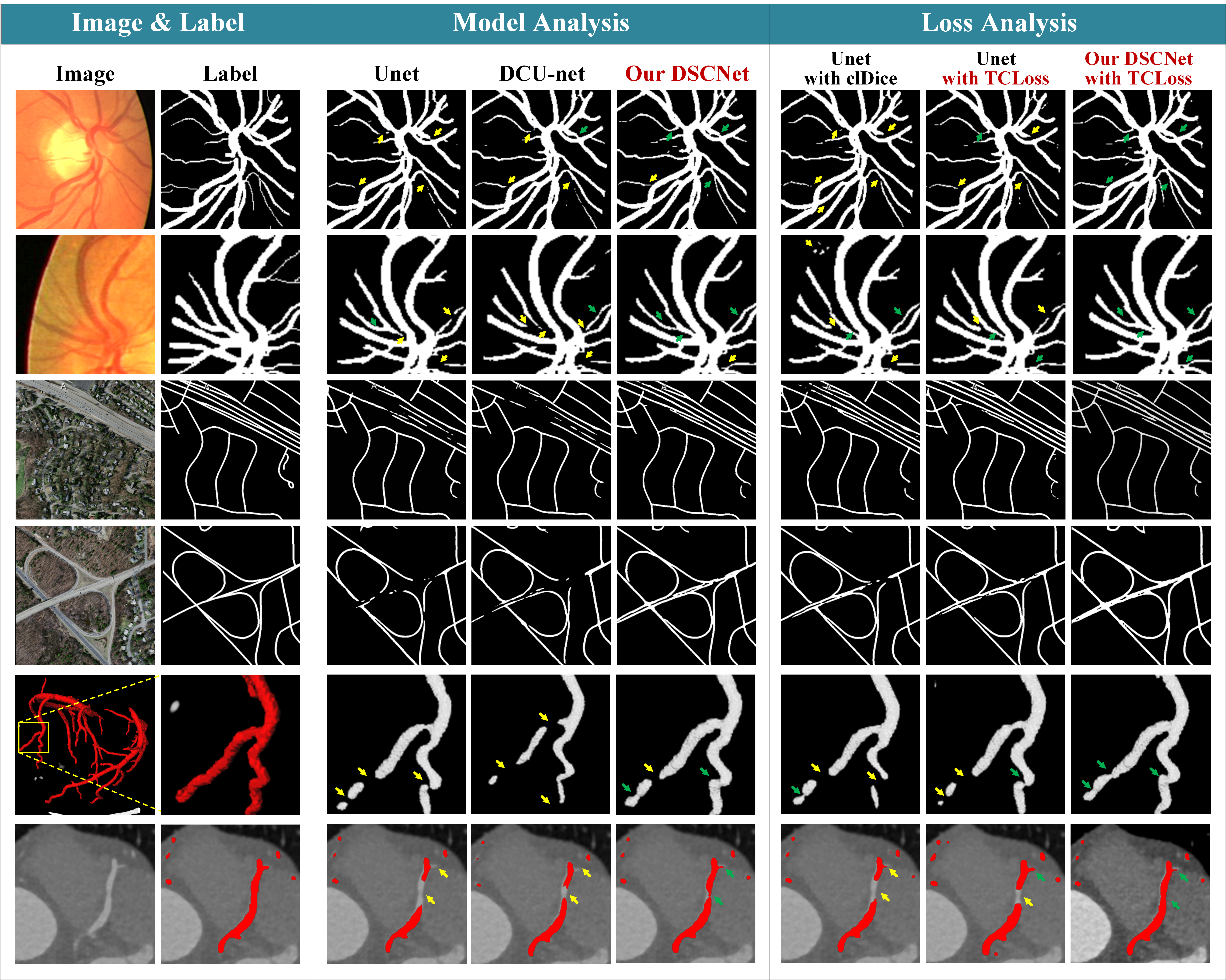}
   \caption{
   \textbf{Qualitative results.} To verify the performance of our method more objectively and efficiently, we selected representative hard-to-segment regions from each dataset. From top to bottom, we show two rows of results for the DRIVE dataset, the Massachusetts road dataset, and our internal Coronary dataset. From left to right, we show the original image, groundtruth, and the results from classical UNet, DCU-net, our DSCNet, UNet with clDice, UNet with our proposed TCLoss, and our DSCNet with TCLoss. The results indicate that our DSCNet and TCLoss outperform the other models regarding segmentation accuracy and topological continuity. The yellow arrows indicate the areas where the segmentation is broken, while the green arrows indicate areas where the segmentation is performing well.
   }
   \label{fig:Results}
\end{figure*}

%------------------------------Table2D----------------------------------
\begin{table*}[!htb]
\begin{center}
\resizebox{\textwidth}{!}{
  \begin{tabular}{ccc|ccccc|cc|c}

\toprule   
\multirow{2}{*}{Dataset} &\multirow{2}{*}{Network} &\multirow{2}{*}{Loss} 
&\multicolumn{5}{|c|}{Volumetric (\%) $\uparrow$} &\multicolumn{2}{|c|}{Topology $\downarrow$} &Distance $\downarrow$\\
\cline{4-11}
 & & &Dice &RDice &clDice &ACC &AUC &$\beta_0$ &$\beta_1$ &HD\\
    
\midrule

%-------------------------------DRIVE------------------------------------
\multirow{12}{*}{DRIVE}		
             				            				
&UNet					&$\mathcal{L}_{CE}$ 
&80.73$_{\pm 1.77}$	&87.94$_{\pm 3.32}$	&79.66$_{\pm 4.00}$	&96.74$_{\pm 0.28}$
&88.57$_{\pm 2.44}$	&1.209$_{\pm 0.342}$	&0.883$_{\pm 0.135}$	&6.86$_{\pm 0.56}$\\

&Transunet				&$\mathcal{L}_{CE}$ 
&80.56$_{\pm 2.14}$	&87.14$_{\pm 3.82}$	&79.02$_{\pm 5.05}$	&96.75$_{\pm 0.32}$
&88.02$_{\pm 2.79}$	&1.210$_{\pm 0.309}$	&0.844$_{\pm 0.157}$	&6.83$_{\pm 0.52}$ \\

&CS$^2$-Net				&$\mathcal{L}_{CE}$ 
&77.53$_{\pm 2.94}$	&82.55$_{\pm 4.10}$	&74.88$_{\pm 5.27}$	&96.46$_{\pm 0.36}$
&84.73$_{\pm 2.82}$	&1.391$_{\pm 0.331}$	&0.906$_{\pm 0.177}$	&6.90$_{\pm 0.48}$ \\

&DCU-net				&$\mathcal{L}_{CE}$ 
&80.83$_{\pm 1.99}$	&87.73$_{\pm 3.60}$	&80.19$_{\pm 4.80}$	&96.77$_{\pm 0.31}$	
&88.45$_{\pm 2.67}$	&1.104$_{\pm 0.327}$	&0.817$_{\pm 0.166}$	&6.84$_{\pm 0.58}$ \\

&\textbf{DSCNet(ours)}	&$\mathcal{L}_{CE}$ 
&\textbf{81.85$_{\pm 1.74}$}	&\textbf{88.93$_{\pm 3.36}$}	
&\textbf{81.16$_{\pm 4.54}$}	&\textbf{96.91}$_{\pm 0.28}$
&\textbf{89.38$_{\pm 2.54}$}	&\textbf{1.094$_{\pm 0.301}$}	
&\textbf{0.780$_{\pm 0.162}$}	&\textbf{6.68$_{\pm 0.49}$} \\

\cline{2-11}
		            				
&UNet					&\textbf{$\mathcal{L}_{TC}$(ours)} 
&80.93$_{\pm 1.97}$	&88.00$_{\pm 3.41}$	&80.28$_{\pm 4.41}$	&96.78$_{\pm 0.30}$	
&88.63$_{\pm 2.56}$ &1.117$_{\pm 0.286}$	&\textbf{0.797$_{\pm 0.151}$}	&6.88$_{\pm 0.53}$ \\

&Transunet				&\textbf{$\mathcal{L}_{TC}$(ours)} 
&80.79$_{\pm 2.11}$	&87.78$_{\pm 3.80}$	&79.86$_{\pm 4.90}$	&96.76$_{\pm 0.32}$	
&88.48$_{\pm 2.82}$	&1.176$_{\pm 0.295}$	&0.818$_{\pm 0.176}$	&6.83$_{\pm 0.51}$ \\

&CS$^2$-Net				&\textbf{$\mathcal{L}_{TC}$(ours)} 
&79.69$_{\pm 2.31}$	&86.14$_{\pm 3.82}$	&77.72$_{\pm 5.09}$	&96.64$_{\pm 0.32}$	
&87.25$_{\pm 2.76}$	&1.308$_{\pm 0.334}$	&0.848$_{\pm 0.160}$	&6.93$_{\pm 0.45}$ \\

&DCU-net				&\textbf{$\mathcal{L}_{TC}$(ours)} 
&81.18$_{\pm 1.90}$	&87.89$_{\pm 3.43}$	&80.60$_{\pm 4.54}$	&96.83$_{\pm 0.31}$
&88.59$_{\pm 2.57}$	&1.076$_{\pm 0.313}$	&0.817$_{\pm 0.167}$	&6.80$_{\pm 0.56}$ \\

&UNet					&clDice 
&80.77$_{\pm 1.92}$	&87.53$_{\pm 3.42}$	&79.93$_{\pm 4.48}$	&96.77$_{\pm 0.31}$	
&88.29$_{\pm 2.52}$ &1.199$_{\pm 0.303}$	&0.833$_{\pm 0.157}$	&6.93$_{\pm 0.54}$ \\
	
%&UNet					&DDT 
%&80.96$_{\pm 2.06}$	&87.62$_{\pm 3.66}$	&79.71$_{\pm 4.79}$	&96.81$_{\pm 0.33}$	
%&88.37$_{\pm 2.70}$	&1.204$_{\pm 0.350}$	&0.816$_{\pm 0.147}$	&6.80$_{\pm 0.54}$ \\

&UNet					&$\mathcal{L}_{WTC}$
&80.89$_{\pm 1.95}$	&87.85$_{\pm 3.55}$	&80.03$_{\pm 4.75}$	&96.78$_{\pm 0.29}$	
&88.53$_{\pm 2.64}$	&1.144$_{\pm 0.339}$	&0.814$_{\pm 0.176}$	&6.79$_{\pm 0.47}$ \\

&\textbf{DSCNet(ours)}	&\textbf{$\mathcal{L}_{TC}$(ours)} 
&\textbf{82.06$_{\pm 1.44}$}	&\textbf{90.17$_{\pm 3.04}$}	
&\textbf{82.07$_{\pm 4.35}$}	&\textbf{96.87$_{\pm 0.24}$}
&\textbf{90.27$_{\pm 2.32}$}	&\textbf{0.998$_{\pm 0.312}$}	&0.803$_{\pm 0.179}$	
&\textbf{6.78$_{\pm 0.51}$} \\
\bottomrule
%-------------------------------Roads------------------------------------
\toprule 
\multirow{6}{*}{ROADS}		
            				            				
&UNet					&$\mathcal{L}_{CE}$ 
&76.90$_{\pm 6.30}$	&84.07$_{\pm 6.46}$	&86.87$_{\pm 6.59}$	
&97.97$_{\pm 1.27}$	&98.29$_{\pm 1.24}$	&1.107$_{\pm 0.551}$	&1.505$_{\pm 0.467}$	&8.11$_{\pm 2.42}$ \\

&Transunet				&$\mathcal{L}_{CE}$ 
&75.82$_{\pm 6.83}$	&81.50$_{\pm 6.65}$	&86.04$_{\pm 7.40}$	
&97.97$_{\pm 1.28}$	&98.23$_{\pm 1.15}$	&1.105$_{\pm 0.615}$	&1.570$_{\pm 0.663}$	&8.11$_{\pm 2.53}$ \\

&DCU-net				&$\mathcal{L}_{CE}$ 
&77.24$_{\pm 6.30}$	&84.26$_{\pm 6.37}$	&86.98$_{\pm 6.53}$	
&98.03$_{\pm 1.14}$	&98.34$_{\pm 1.19}$	&1.085$_{\pm 0.653}$		
&1.474$_{\pm 0.497}$		&8.04$_{\pm 2.53}$ \\

%\cline{2-11}
		            				
&UNet					&\textbf{$\mathcal{L}_{TC}$(ours)} 
&77.70$_{\pm 6.07}$	&84.80$_{\pm 5.96}$	&87.47$_{\pm 6.31}$	
&98.03$_{\pm 1.23}$	&98.41$_{\pm 1.13}$	&1.072$_{\pm 0.631}$	&1.401$_{\pm 0.496}$&8.04$_{\pm 2.72}$	 \\

&UNet					&clDice 
&77.37$_{\pm 5.57}$	&84.18$_{\pm 5.99}$	&87.05$_{\pm 6.34}$	&98.03$_{\pm 1.22}$	
&98.40$_{\pm 1.12}$	&1.079$_{\pm 0.613}$	
&1.407$_{\pm 0.603}$&8.08$_{\pm 2.46}$ \\

%&UNet					&$\mathcal{L}_{WTC}$
%&	&	&	&
%&	&	&	& \\

&\textbf{DSCNet(ours)}	&$\mathcal{L}_{CE}$ 
&78.04$_{\pm 5.77}$	&85.35$_{\pm 5.42}$	&\textbf{87.74$_{\pm 6.02}$}	
&98.05$_{\pm 1.21}$ &98.39$_{\pm 1.19}$	&1.118$_{\pm 0.641}$	&1.441$_{\pm 0.523}$	&7.96$_{\pm 2.43}$ \\
	
&\textbf{DSCNet(ours)}	&\textbf{$\mathcal{L}_{TC}$(ours)} 
&\textbf{78.21$_{\pm 5.77}$}	&\textbf{85.85$_{\pm 5.56}$}	&87.64$_{\pm 5.99}$	
&\textbf{98.05$_{\pm 1.21}$}
&\textbf{98.46$_{\pm 1.08}$}	&\textbf{1.053$_{\pm 0.523}$}	&\textbf{1.396$_{\pm 0.456}$}	&\textbf{7.34$_{\pm 2.48}$} \\
\bottomrule
\end{tabular}
}
\end{center}
  \caption{Quantitative results for the DRIVE retina dataset and the Massachusetts road dataset (ROADS) are shown in this table. All experiments verified the performance of our method on three levels: volumetric accuracy, topological connectivity, and distance error. Our DSCNet and TCLoss achieve the most competitive results in all metrics.
}  

  \label{tab:results}
\end{table*}

%-----------------------------Table3D----------------------------------
\begin{table*}[!htb]
\begin{center}
\resizebox{\textwidth}{!}{
\begin{tabular}{ccc|ccc|ccc|c}
\toprule   
\multirow{2}{*}{Dataset} &\multirow{2}{*}{Network} &\multirow{2}{*}{Loss} 
&\multicolumn{3}{|c|}{Volumetric (\%) $\uparrow$} &\multicolumn{3}{|c|}{Topology  OF $\uparrow$} &Distance $\downarrow$\\
\cline{4-10}
 & & &Dice &RDice &clDice &LAD &LCX &RCA &HD \\
    
\midrule
%----------------------------Coronary------------------------------------
\multirow{6}{*}{CORONARY}		
             				            				
&UNet					&$\mathcal{L}_{CE}$ 
&76.87$_{\pm 5.38}$	&84.48$_{\pm 4.55}$	&81.43$_{\pm 6.02}$	
&0.806$_{\pm 0.252}$&0.847$_{\pm 0.239}$	&0.849$_{\pm 0.267}$	&7.727$_{\pm 3.30}$\\

&Transunet				&$\mathcal{L}_{CE}$ 
&76.70$_{\pm 6.65}$	&83.23$_{\pm 6.72}$	&78.71$_{\pm 6.93}$	
&0.810$_{\pm 0.274}$&0.694$_{\pm 0.307}$	&0.816$_{\pm 0.303}$	&8.580$_{\pm 4.11}$\\

&DCU-net				&$\mathcal{L}_{CE}$ 
&78.33$_{\pm 5.00}$	&85.67$_{\pm 4.29}$	&82.29$_{\pm 5.31}$	
&0.833$_{\pm 0.219}$&0.746$_{\pm 0.296}$	&0.835$_{\pm 0.300}$	&7.331$_{\pm 3.06}$ \\

%\cline{2-11}
		            				
%&UNet					&\textbf{$\mathcal{L}_{TC}$(ours)} 
%&78.42	&84.26	&82.53	&	&&&6.761$_{\pm 3.38}$	 \\

&UNet					&clDice 
&77.86$_{\pm 5.25}$	&84.42$_{\pm 4.65}$	&82.37$_{\pm 5.54}$	
&0.817$_{\pm 0.256}$&0.845$_{\pm 0.234}$	&0.859$_{\pm 0.265}$	&7.412$_{\pm 3.68}$ \\
	
&\textbf{DSCNet(ours)}	&$\mathcal{L}_{CE}$ 
&79.92$_{\pm 5.26}$	&85.98$_{\pm 4.60}$	&84.95$_{\pm 5.76}$	
&0.858$_{\pm 0.198}$&0.853$_{\pm 0.241}$	&0.862$_{\pm 0.267}$	&6.326$_{\pm 2.85}$\\	

&\textbf{DSCNet(ours)}	&\textbf{$\mathcal{L}_{TC}$(ours)}
&\textbf{80.27$_{\pm 4.67}$}	&\textbf{86.37$_{\pm 4.16}$}	
&\textbf{85.26$_{\pm 4.98}$}
&\textbf{0.866$_{\pm 0.195}$}	&\textbf{0.885$_{\pm 0.210}$}	&\textbf{0.882$_{\pm 0.250}$}	&\textbf{5.787$_{\pm 2.99}$} \\
\bottomrule
\end{tabular}
}
\end{center}
  \caption{Quantitative results for the 3D Cardiac CCTA dataset. Experimental metrics supplement with OF metrics for verifying the topological connectivity. The LAD, LCX, and RCA are the three main trunks of the coronary vessels and are of great clinical value.}
  \label{tab:results2}
\end{table*}

In this paragraph, we will evaluate and analyze our proposed framework's effectiveness in three ways: (1) The performance of our proposed method for the thin tubular structure segmentation task is compared and verified by the following metrics. The visual effects of different methods are simultaneously shown. (2) We analyzed the effectiveness of our proposed DSConv to guide the model to focus on the tubular structure, and the help of the TCLoss to constrain the topology of the segmentation. (3) We provide comprehensive experiments, including ablation studies, on the DRIVE dataset as an example. Additionally, due to space constraints, we highlight some of the most important comparison experiments on other datasets. The results show the strong performance of our method on both 2D and 3D fields.

%------------------------------------------------------------------------
\subsection{Quantitative Evaluation}
The advantages of our method on each metric are demonstrated in Table~\ref{tab:results}, and the results show that our proposed DSCNet achieves better results on both 2D and 3D datasets. 

\textbf{Evaluation on DRIVE.}
On the DRIVE dataset, our DSCNet outperforms the other models regarding segmentation accuracy and topological continuity. In Table~\ref{tab:results}, our proposed DSCNet achieves the best segmentation results compared with other methods with Dice of 82.06\%, RDice of 90.17\%, clDice of 82.07\%, ACC of 96.87\%, and AUC of 90.27\% from the perspective of the volumetric accuracy. Meanwhile, from the view of the topology, our DSCNet achieves the best topological continuity compared with other methods with $\beta_0$ error of 0.998 and $\beta_1$ error of 0.803. The results show that our method better captures the specific features of the thin tubular structures and exhibits a more accurate segmentation performance and a more continuous topology. As shown in the sixth to twelfth rows of Table~\ref{tab:results}, with the addition of our TCLoss, different models both show improvement in the topological continuity of the segmentation. The results illustrate that our TCLoss accurately constrains the model to focus on thin tubular structures that lose topological continuity.

\textbf{Evaluation on ROADS.}
On the Massachusetts Roads dataset, our DSCNet also achieves the best results. As shown in Table~\ref{tab:results}, our proposed DSCNet with TCLoss achieves the best segmentation results compared with other methods with Dice of 78.21\%, RDice of 85.85\%, and clDice of 87.64\%. Compared with the results of the classical segmentation network UNet, our method achieves at most 1.31\% Dice, 1.78\% RDice, and 0.77\% clDice improvements. The results show that our model also performs well for structurally complex and morphologically variable road datasets compared with other models.

\textbf{Evaluation on CORONARY.}
On the Cardiac CCTA dataset, we verify that our DSCNet still achieves the same best results for segmenting the thin tubular structures in 3D. As shown in Table~\ref{tab:results2}, our proposed DSCNet achieves the best segmentation results compared with other methods with Dice of 80.27\%, RDice of 86.37\%, and clDice of 85.26\%. Compared with the results of the classical segmentation network UNet, our method achieves at most 3.40\% Dice, 1.89\% RDice, and 3.83\% clDice improvements. Meanwhile, we used OF metrics to assess the continuity of segmentation. With our method, the OF metrics improved by 6.00\% for LAD, 3.78\% for LCX, and 3.30\% for RCA (LAD, LCX, and RCA are the vital trunks of the coronary vessels). The improvement in the continuity of vessels plays a crucial role in the clinic.

%Meanwhile, our DSCNet achieves the best topological continuity on the three main trunks with 85.75\% OF of LAD, 85.30\% OF of LCX, and 86.20\% OF of RCA improvements. 

\textbf{Ablation Experiment Analysis.}
Taking the DRIVE dataset as an example, the ablation experiments prove the importance of our DSCNet and our TCLoss. (1) To prove the effectiveness of our DSCNet. The results in the first five rows of Table~\ref{tab:results} show that our method is better suited to segmenting thin tubular structures. The results show that our proposed DSConv plays a vital role in the model, helping the network better to capture the critical features of the thin tubular structures. (2) To prove the effectiveness of our TCLoss. As shown in the sixth to ninth rows of Table~\ref{tab:results}, with the addition of our TCLoss, different models both show improvement in the topological continuity of the segmentation. The results illustrate that our TCLoss accurately constrains the model to focus on thin tubular structures that lose topological continuity. 

%------------------------------------------------------------------------
\subsection{Qualitative Evaluation}
Our DSCNet and TCLoss have decisive visual superiority in arbitrary aspects (Figure~\ref{fig:Results}). (1) To demonstrate the effectiveness of our DSCNet. From left to right, the third to fifth columns show the performance of the different networks in terms of segmentation accuracy. Thanks to our DSConv to adaptively perceive critical features of the thin tubular structure, our model focuses more accurately on special tubular features than other methods, thus showing better performance on tubular structure segmentation. (2) To demonstrate the effectiveness of our TCLoss. From left to right, the sixth to eighth columns show the performance of different loss functions on the continuity of the segmentation of the thin tubular structure. With the addition of our proposed TCLoss, the continuity of the segmentation is greatly improved in hard-to-segment regions. The results confirm that our method gives a stable segmentation performance with better topological continuity, especially in complex and variable morphological structures. Notably, on the Massachusetts Roads dataset, our model achieves good visualization on adjacent straight or curved roads. More visualization results can be found in the supplementary material.

%---------------------------------------------------------------------
\subsection{Model Analysis}
\begin{figure}[t]
  \centering
%  \fbox{\rule{0pt}{2in} \rule{0.9\linewidth}{0pt}}
   \includegraphics[width=\linewidth]{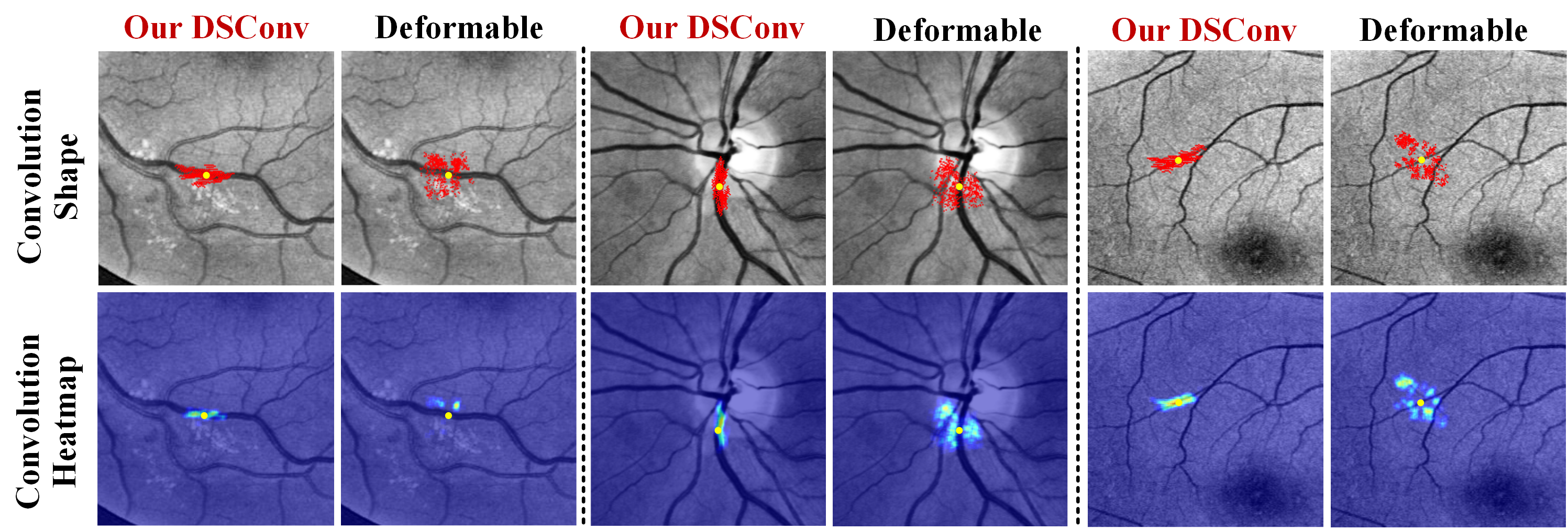}
   \caption{\textbf{Top}: We superimpose 3 layers with a total of 729 points (red) on each image to show the convolution kernel’s position and shape for a given point (yellow). \textbf{Bottom}: The heatmap shows the region of interest for each convolution.}
   \label{fig:Model_analysis}
\end{figure}

Our DSConv dynamically adapts the shape to tubular structures, and the attention well fits the target. (1) Adapt to the shape of tubular structures. The top of Figure~\ref{fig:Model_analysis} shows the convolution kernel’s positions and shape. Visualization results show that our DSConv adapts well to tubular structures and maintains the shape, while the deformable convolution wanders outside the target. (2) Focus on the locations of tubular structures. The bottom of Figure~\ref{fig:Model_analysis} shows the heatmap of the attention on the given point. Results show that the brightest regions of our DSConv are concentrated in the tubular structures, which represents that our DSConv is more sensitive to tubular structures.

%------------------------------------------------------------------------
\subsection{Future Work}
Our proposed framework copes well with the segmentation of thin tubular structures and successfully integrates morphological features with topological knowledge to guide the model to adapt to the segmentation. However, whether other morphological targets will achieve better performance with a similar paradigm is still an exciting topic. Meanwhile, more research will investigate the possibility of incorporating other types of domain knowledge or topological analysis to further improve the performance of the segmentation. Furthermore, more experimental and theoretical validation will enrich this topic.

%%%%%%%%% CONCLUSION
\section{Conclusion}
In this study, we focus on the special features of the tubular structures and use this knowledge to guide the model to enhance the perception simultaneously in three stages: feature extraction, feature fusion, and loss constraint. Firstly, we propose a dynamic snake convolution to adaptively focus on the thin and tortuous structure, thus accurately capturing the features of tubular structures. Secondly, we introduce a multi-view feature fusion strategy to complement the focus on features from multiple angles during feature fusion, ensuring the retention of important information from different global morphology. Lastly, we propose a topological continuity constraint loss to constrain the topological continuity of the segmentation. Our method is verified on both 2D and 3D datasets and the results show our method provides better accuracy and continuity on the tubular structure segmentation task compared with several methods.

\section{Acknowledgments}
This research was supported by the Intergovernmental Cooperation Project of the National Key Research and Development Program of China (2022YFE0116700) and the Postgraduate Research $\&$ Practice Innovation Program of Jiangsu Province, the Fundamental Research Funds for the Central Universities (KYCX22$\_$0239). We thank the Big Data Computing Center of Southeast University for providing the facility support.

{\small
\bibliographystyle{ieee_fullname}
\bibliography{egpaper_final}

\begin{thebibliography}{10}\itemsep=-1pt

\bibitem{RW_Loss_Topological}
R.~J. Ara{\'{u}}jo, Jaime~S Cardoso, and H.~P Oliveira.
\newblock {Topological Similarity Index and Loss Function for Blood Vessel
  Segmentation}.
\newblock {\em IEEE TRANSACTIONS ON MEDICAL IMAGING}, XX:1, jul 2021.

\bibitem{RISPER2}
Ulrich Bauer.
\newblock Ripser: efficient computation of {V}ietoris-{R}ips persistence
  barcodes.
\newblock {\em J. Appl. Comput. Topol.}, 5(3):391--423, 2021.

\bibitem{Exp_transunet}
Jieneng Chen, Yongyi Lu, Qihang Yu, et~al.
\newblock Transunet: Transformers make strong encoders for medical image
  segmentation.
\newblock {\em arXiv preprint arXiv:2102.04306}, 2021.

\bibitem{Exp_Unet}
{\"O}zg{\"u}n {\c{C}}i{\c{c}}ek, Ahmed Abdulkadir, Soeren~S. Lienkamp, et~al.
\newblock 3d u-net: Learning dense volumetric segmentation from sparse
  annotation.
\newblock In {\em International Conference on Medical Image Computing and
  Computer-Assisted Intervention}, pages 424--432. Springer, 2016.

\bibitem{PH_3}
David Cohen-Steiner, Herbert Edelsbrunner, John Harer, et~al.
\newblock Lipschitz functions have l p-stable persistence.
\newblock {\em Foundations of computational mathematics}, 10(2):127--139, 2010.

\bibitem{RW_DCN}
Jifeng Dai, Haozhi Qi, Yuwen Xiong, et~al.
\newblock Deformable convolutional networks.
\newblock In {\em Proceedings of the IEEE International Conference on Computer
  Vision (ICCV)}, Oct 2017.

\bibitem{RW_DCN_CMR}
Shunjie Dong, Zixuan Pan, Yu Fu, et~al.
\newblock Deu-net 2.0: Enhanced deformable u-net for 3d cardiac cine mri
  segmentation.
\newblock {\em Medical Image Analysis}, 78:102389, 2022.

\bibitem{RW_Net_wavelet}
Yuting He, Rongjun Ge, Jiasong Wu, et~al.
\newblock Thin semantics enhancement via high-frequency priori rule for thin
  structures segmentation.
\newblock In {\em 2021 IEEE International Conference on Data Mining (ICDM)},
  pages 1096--1101. IEEE, 2021.

\bibitem{PH_1}
Edelsbrunner Herbert and Harer John.
\newblock Computational topology: an introduction.
\newblock {\em American Mathematical Soc}, 2010.

\bibitem{PH_WD}
Xiaoling Hu, Fuxin Li, Dimitris Samaras, et~al.
\newblock Topology-preserving deep image segmentation.
\newblock {\em Advances in neural information processing systems}, 32, 2019.

\bibitem{Hausdorff_1}
Daniel~P Huttenlocher, Gregory~A. Klanderman, and William~J Rucklidge.
\newblock Comparing images using the hausdorff distance.
\newblock {\em IEEE Transactions on pattern analysis and machine intelligence},
  15(9):850--863, 1993.

\bibitem{RW_DCN_retinal1}
Qiangguo Jin, Zhaopeng Meng, Tuan~D. Pham, et~al.
\newblock Dunet: A deformable network for retinal vessel segmentation.
\newblock {\em Knowledge-Based Systems}, 178:149--162, 2019.

\bibitem{Intro01KoJACC}
Brian~S. Ko, James~D. Cameron, Ravi~K. Munnur, et~al.
\newblock Noninvasive ct-derived ffr based on~structural and~fluid analysis.
\newblock 10(6):663--673, 2017.

\bibitem{RW_Network_TreeLSTM}
Bin Kong, Xin Wang, Junjie Bai, et~al.
\newblock Learning tree-structured representation for 3d coronary artery
  segmentation.
\newblock {\em Computerized Medical Imaging and Graphics}, 80:101688, 2020.

\bibitem{RW_FF_global_local}
Yang Li, Yue Zhang, Jing-Yu Liu, et~al.
\newblock Global transformer and dual local attention network via deep-shallow
  hierarchical feature fusion for retinal vessel segmentation.
\newblock {\em IEEE Transactions on Cybernetics}, pages 1--14, 2022.

\bibitem{Intro02MinJAMA}
James~K. Min, Jonathon Leipsic, Michael~J. Pencina, et~al.
\newblock {Diagnostic Accuracy of Fractional Flow Reserve From Anatomic CT
  Angiography}.
\newblock {\em JAMA}, 308(12):1237--1245, 09 2012.

\bibitem{Exp_road_2D}
Volodymyr Mnih.
\newblock {\em Machine learning for aerial image labeling}.
\newblock University of Toronto (Canada), 2013.

\bibitem{Exp_CS2}
Lei Mou, Yitian Zhao, Huazhu Fu, et~al.
\newblock Cs2-net: Deep learning segmentation of curvilinear structures in
  medical imaging.
\newblock {\em Medical Image Analysis}, 67:101874, 2021.

\bibitem{PH_2}
James~R Munkres.
\newblock {\em Elements of algebraic topology}.
\newblock CRC press, 2018.

\bibitem{RW_FF_mvsgan}
Xiaoming Qi, Yuting He, Guanyu Yang, et~al.
\newblock Mvsgan: Spatial-aware multi-view cmr fusion for accurate 3d left
  ventricular myocardium segmentation.
\newblock {\em IEEE Journal of Biomedical and Health Informatics},
  26(5):2264--2275, 2021.

\bibitem{RW_FF_CMR}
Xiaoming Qi, Guanyu Yang, Yuting He, et~al.
\newblock Contrastive re-localization and history distillation in federated cmr
  segmentation.
\newblock {\em International Conference on Medical Image Computing and
  Computer-Assisted Intervention}, pages 256--265, 2022.

\bibitem{RW_EE}
Yaolei Qi, Han Xu, Yuting He, et~al.
\newblock Examinee-examiner network: Weakly supervised accurate coronary lumen
  segmentation using centerline constraint.
\newblock {\em IEEE Transactions on Image Processing}, 30:9429--9441, 2021.

\bibitem{OF}
Michiel Schaap, Coert~T Metz, Theo van Walsum, et~al.
\newblock Standardized evaluation methodology and reference database for
  evaluating coronary artery centerline extraction algorithms.
\newblock {\em Medical Image Analysis}, 13(5):701--714, 2009.

\bibitem{RW_Loss_clDice}
Suprosanna Shit, Johannes~C. Paetzold, Anjany Sekuboyina, et~al.
\newblock cldice - a novel topology-preserving loss function for tubular
  structure segmentation.
\newblock In {\em Proceedings of the IEEE/CVF Conference on Computer Vision and
  Pattern Recognition (CVPR)}, pages 16560--16569, June 2021.

\bibitem{Exp_vessel_2d}
Joes Staal, Michael~D Abr{\`a}moff, Meindert Niemeijer, et~al.
\newblock Ridge-based vessel segmentation in color images of the retina.
\newblock {\em IEEE transactions on medical imaging}, 23(4):501--509, 2004.

\bibitem{HD}
Abdel~Aziz Taha and Allan Hanbury.
\newblock Metrics for evaluating 3d medical image segmentation: analysis,
  selection, and tool.
\newblock {\em BMC medical imaging}, 15(1):1--28, 2015.

\bibitem{RISPER1}
Christopher Tralie, Nathaniel Saul, and Rann Bar-On.
\newblock {Ripser.py}: A lean persistent homology library for python.
\newblock {\em The Journal of Open Source Software}, 3(29):925, Sep 2018.

\bibitem{RW_Network_Points}
Dong Wang, Zhao Zhang, Ziwei Zhao, et~al.
\newblock Pointscatter: Point set representation for tubular structure
  extraction, 2022.

\bibitem{RW_Loss_DDT}
Yan Wang, Xu Wei, Fengze Liu, et~al.
\newblock Deep distance transform for tubular structure segmentation in ct
  scans.
\newblock In {\em Proceedings of the IEEE/CVF Conference on Computer Vision and
  Pattern Recognition (CVPR)}, June 2020.

\bibitem{RW_Loss_PH}
Chi-Chong Wong and Chi-Man Vong.
\newblock Persistent homology based graph convolution network for fine-grained
  3d shape segmentation.
\newblock In {\em Proceedings of the IEEE/CVF International Conference on
  Computer Vision (ICCV)}, pages 7098--7107, October 2021.

\bibitem{RW_DCN_retinal2}
Xin Yang, Zhiqiang Li, Yingqing Guo, et~al.
\newblock {DCU-net: a deformable convolutional neural network based on cascade
  U-net for retinal vessel segmentation}.
\newblock {\em Multimedia Tools and Applications}, 81(11):15593--15607, may
  2022.

\bibitem{RW_Dilated}
Fisher Yu, Vladlen Koltun, and Thomas Funkhouser.
\newblock Dilated residual networks.
\newblock In {\em Proceedings of the IEEE Conference on Computer Vision and
  Pattern Recognition (CVPR)}, July 2017.

\bibitem{RW_FF_MICCAI}
Xiao Zhang, Jingyang Zhang, Lei Ma, et~al.
\newblock Progressive deep segmentation of coronary artery via hierarchical
  topology learning.
\newblock In {\em Medical Image Computing and Computer Assisted Intervention --
  MICCAI 2022}, pages 391--400, Cham, 2022. Springer Nature Switzerland.

\bibitem{RW_TIP_superpixel}
Chunhui Zhao, Wenxiang Zhu, and Shou Feng.
\newblock Superpixel guided deformable convolution network for hyperspectral
  image classification.
\newblock {\em IEEE Transactions on Image Processing}, 31:3838--3851, 2022.

\bibitem{RW_FF_graph}
Gangming Zhao, Kongming Liang, Chengwei Pan, et~al.
\newblock {Graph Convolution Based Cross-Network Multi-Scale Feature Fusion for
  Deep Vessel Segmentation}.
\newblock {\em IEEE Transactions on Medical Imaging}, pages 1--1, 2022.

\end{thebibliography}
}

\end{document}